\documentclass{CUP-JNL-DCE}

\usepackage{latexsym}
\usepackage{graphicx}
\usepackage{multicol,multirow}
\usepackage{amsmath,amssymb,amsfonts}
\usepackage{mathrsfs}
\usepackage{amsthm}
\usepackage{rotating}
\usepackage{appendix}
\usepackage[authoryear]{natbib}
\usepackage{ifpdf}
\usepackage[T1]{fontenc}
\usepackage{times}
\usepackage{sourcesanspro}
\usepackage{newtxmath}
\usepackage{textcomp}%
\usepackage{xcolor}%
\usepackage{hyperref}
\usepackage{subcaption}
\articletype{RESEARCH ARTICLE}
\jname{Pre-print submitted to Data-Centric Engineering}
\begin{document}

\begin{Frontmatter}

\title{Deep learning modelling of manufacturing and build variations on multi-stage axial compressors aerodynamics}

\author*[1,2]{Giuseppe Bruni}\orcid{0000-0002-8514-4315}\email{giuseppe.bruni@siemens-energy.com}
\author[2]{Sepehr Maleki}\orcid{0000-0001-6897-7385}
\author[1,2]{Senthil K. Krishnababu}\orcid{0009-0003-2745-8783}

\authormark{Bruni \textit{et al.}}

\address*[1]{\orgname{Lincoln AI Lab, University of Lincoln}, \orgaddress{\street{Brayford Way}, \state{Lincoln}, \postcode{LN6 7DL}, \country{United Kingdom}}}
\address[2]{\orgname{Siemens Energy}, \orgaddress{\street{1 Waterside South}, \state{Lincoln}, \postcode{LN5 7FD}, \country{United Kingdom}}}

\received{}
\revised{}
\accepted{}


\abstract{Applications of deep learning to physical simulations such as Computational Fluid Dynamics have recently experienced a surge in interest, and their viability has been demonstrated in different domains. However, due to the highly complex, turbulent and three-dimensional flows, they have not yet been proven usable for turbomachinery applications. Multi-stage axial compressors for gas turbine applications represent a remarkably challenging case, due to the high-dimensionality of the regression of the flow-field from geometrical and operational variables. This paper demonstrates the development and application of a deep learning framework for predictions of the flow field and aerodynamic performance of multi-stage axial compressors. A physics-based dimensionality reduction approach unlocks the potential for flow-field predictions, as it re-formulates the regression problem from an un-structured to a structured one, as well as reducing the number of degrees of freedom. Compared to traditional "black-box" surrogate models, it provides explainability to the predictions of the overall performance by identifying the corresponding aerodynamic drivers. The model is applied to manufacturing and build variations, as the associated performance scatter is known to have a significant impact on $CO_2$ emissions, which poses a challenge of great industrial and environmental relevance. The proposed architecture is proven to achieve an accuracy comparable to that of the CFD benchmark, in real-time, for an industrially relevant application. The deployed model, is readily integrated within the manufacturing and build process of gas turbines, thus providing the opportunity to analytically assess the impact on performance with actionable and explainable data.}

\begin{policy}[Impact Statement]
This paper demonstrates the development and application of a deep learning framework for predictions of the flow-field and aerodynamic performance of multi-stage axial compressors. A physics-based dimensionality reduction unlocks the potential to use supervised learning approaches for turbomachinery flow field predictions, while providing explainability to the overall performance predictions. The model is applied to manufacturing and build variations, as the associated scatter in compressor efficiency is known to have a significant impact on the overall performance and $CO_2$ emissions of the gas turbine, thus posing a challenge of great industrial and environmental relevance. The proposed architecture is proven to achieve an accuracy comparable to that of the CFD benchmark. The deployed model, is readily integrated within the manufacturing and build process of gas turbines, thus providing the opportunity to analytically assess the impact on performance with actionable and explainable data, and reduce requirements for expensive physical tests.
\end{policy}

\end{Frontmatter}

\section{Introduction}

The highly complex, turbulent and three-dimensional flow-field of multi-stage axial compressors has traditionally been one the main challenges for the application of CFD (Computational Fluid Dynamics) in the turbomachinery industry. In this context, the flow field consists of the aerodynamic variables of interest (such as pressure, temperature, velocities, etc.) predicted on a numerical discretization of the domain (i.e. mesh or grid), solving the Reynolds-Averaged Navier Stokes equations (RANS). Advancements in modelling and numerical methods have progressively improved accuracy and computational cost, making CFD analyses an integral part of industrial design processes, greatly improving the design capabilities of gas turbine manufacturers. This has led to an interest in the development of robust design strategies to account for real world effects in the design and analyses, which are known to affect the overall performance as well as the operability of the compressor. Gas turbine manufacturers have gathered vast amount of operational data over the years, enabling them to establish best practices and design guidelines for manufacturing and build processes. These guidelines, often consider a trade-off between product cost and engine performance, with tolerance ranges specified accordingly. However, the impact on performance, used to define these ranges, is typically based on previous experience and simplified correlations. The identification of  manufacturing and build uncertainties impact on the operational performance can be used to both drive improvements in the manufacturing technology, and inform the design strategies to produce robust designs that are desensitized to these variabilities.

\subsection{The challenge}
Characterizing the impact of all the relevant manufacturing and build variations which are expected to affect the overall performance of a given compressor design, can be achieved by engine build-specific CFD models. While having demonstrated to provide excellent agreement with test data \cite{Bruni2024CFD}, these models are not traditionally used as part of the manufacturing and build process as a day-to-day occurrence, due to the associated computational cost and requirement for specialized engineers to carry out the analyses. Therefore, there is a technology gap for the development of predictive models that can match the accuracy of CFD, but with quicker turn-around times.

\subsection{Our contribution}
\begin{figure}[!htbp]
   \centering
      \includegraphics[width=\textwidth]{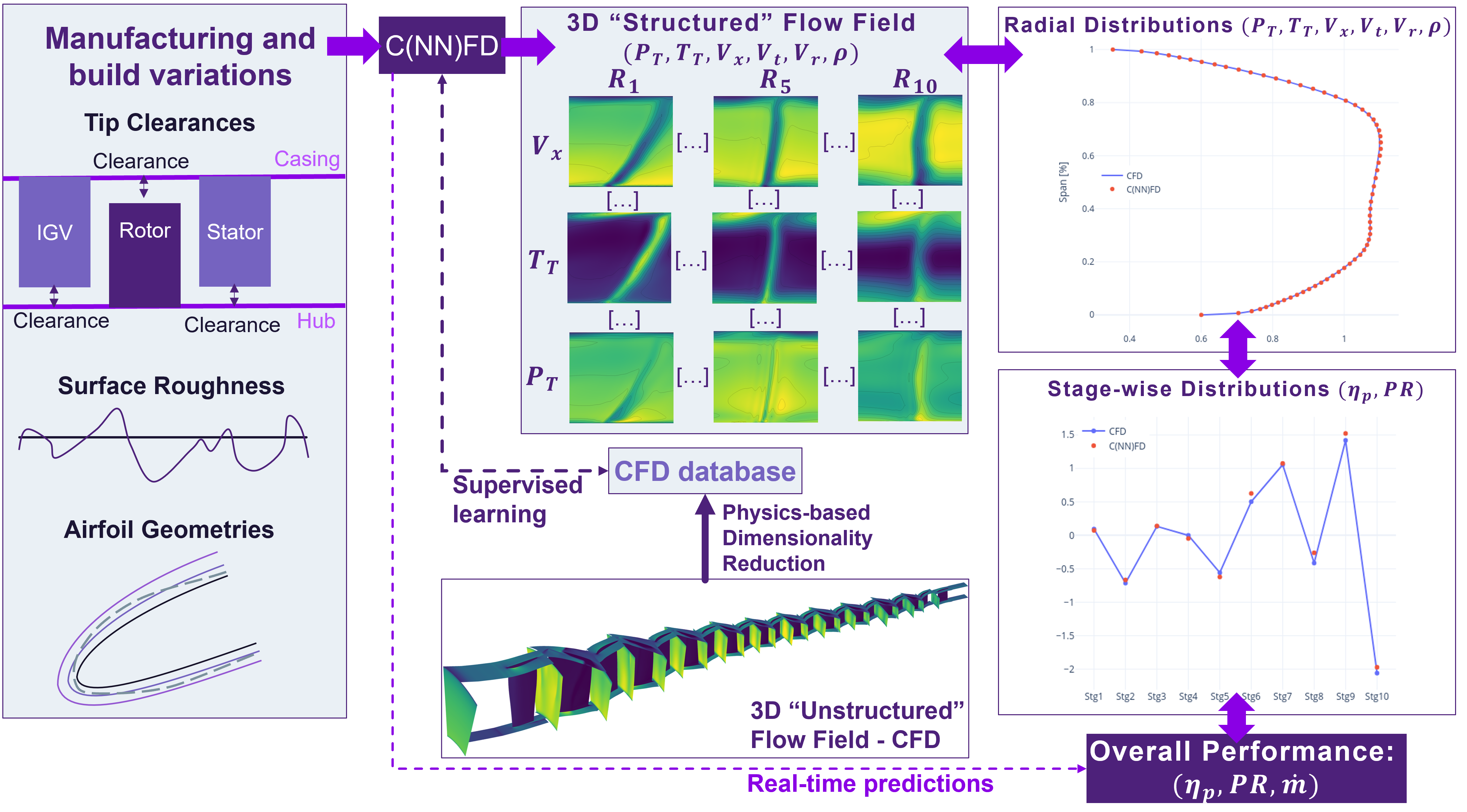}
      \caption{Overview of the C(NN)FD framework}
      \label{fig:graphical_abstract}
\end{figure}

This paper demonstrates the development and application of C(NN)FD, a deep learning framework previously introduced by the authors \cite{Bruni2023} for predictions of the flow field and aerodynamic performance of turbomachinery. The scalability is demonstrated from a proof of concept to a significantly more computationally expensive use-case of great industrial relevance, multi-stage axial compressors for gas turbine applications. We propose a physics-based dimensionality reduction to unlock the potential to use supervised learning approaches for flow-field predictions, as it re-formulates the regression problem from un-structured to structured, as well as reducing the number of degrees of freedom. The proposed architecture also has a competitive advantage compared to traditional "black-box" surrogate models, as it provides explainability to the predictions of variations in the overall performance. The corresponding aerodynamic drivers are identified by providing predictions for the 3D "structured" flow-field, as well as radial distributions at all the inter-row locations, 1D row-wise distributions and stage-wise distributions.  The model is applied to manufacturing and build variations, as the associated scatter in efficiency is known to have a significant impact on the overall performance and $CO_2$ emissions of the gas turbine, therefore posing a challenge of great industrial and environmental relevance. The proposed architecture is proven to achieve an accuracy comparable to that of the CFD benchmark, in real-time, for an industrially relevant application. The deployed model, is readily integrated within the manufacturing and build process of gas turbines, thus providing the opportunity to analytically assess the impact on performance with actionable and explainable data. This unlocks the capability to model each engine build avoiding the high computational costs, long lead times, specialized personnel requirements of traditional CFD analyses, and reducing the requirement for expensive physical tests. An overview of the C(NN)FD framework is provided in \autoref{fig:graphical_abstract}.

\section{Literature Review}
\subsection{Effect of manufacturing and build variations on the aerodynamic performance}

The impact of manufacturing and build variations on the overall performance of gas turbines is known to be significant \cite{Montomoli2018} \cite{Wang2020}, with the axial compressor being a large contributor. For a given design, compressor efficiency is affected by a combination of in-tolerance variations, which can lead to a substantial increase in $CO_2$ emissions compared to the design target. Therefore, accurately predicting the performance of each engine build by modelling the impact of manufacturing and build variations on engine performance within a short timescale is of great industrial and environmental importance. The focus of the current work is on tip clearance and surface roughness variations, two of the main sources of performance variability, but the framework is readily generalisable to other manufacturing variations. A detailed review of the effect of the various manufacturing and build variations on the compressor aerodynamic performance, and the associated modelling strategies has been previously been provided by the authors \cite{Bruni2024CFD}. In the following sections, a brief overview is included to provide the readers with the necessary background to interpret the results presented in this paper.

\subsubsection{Tip Clearance variation background}

The term tip clearance refers to the radial distance between the stationary and rotating components within the compressor, such as the rotor blades and casing, or stator blades and rotor hub. Generally, the tip clearance must be large enough to prevent rubbing during operation, but small enough to avoid excessive aerodynamic losses. Axial compressors are designed to operate at a defined tip clearance, which can vary across stages. During manufacturing, the "cold clearance" is specified and measured as part of the build process in stationary conditions. However, during operation, the "hot clearance" can be significantly smaller due to thermal expansion and centrifugal effects. The cold clearance values have a target value and a tolerance range, set by gas turbine manufacturers, based on trade-offs between undesirable rubbing and minimizing tip leakage losses. For a given "cold clearance", the corresponding "hot clearance" can be calculated using either analytical models or operational experience. The "hot clearance" is the actual clearance found during operation, and therefore is used in the CFD and machine learning models presented in this paper. The aerodynamic mechanisms driving tip leakage flows have been extensively researched over the years \cite{Storer1991} and can be modeled analytically with varying degrees of fidelity. Standard practice in both academia and industry, is to use steady-state single passage CFD calculations to model the effects of tip clearance variations on the overall performance \cite{Sakulkaew2013}, which are used as ground-truth for this study. This assumes that circumferential variations in tip clearance can be neglected. Overall, non-axisymmetric tip clearance variations are known to affect the overall performance and surge margin \cite{Suriyanarayanan2022} and are of relevance especially for in-service degradation. However, for manufacturing variations, they can be neglected \cite{Bruni2023} and single passage models can be used for the analyses.

\subsubsection{Surface roughness variation background}

Traditionally, CFD analyses are performed assuming aerofoil and gas path surfaces to be aerodynamically smooth. However, this assumption often does not hold true for production components, which typically exhibit non-negligible roughness due to coating selection or manufacturing methods. A trade-off exists between production cost and surface roughness, as achieving smoother finishes tends to be progressively more expensive. Therefore, it is crucial to characterize the impact of different surface roughness levels on aerodynamic performance \cite{Bruni2024CFD}. This is particularly relevant when comparing engine test data from different builds with coatings from various suppliers, which can exhibit significantly different roughness levels. Experimental studies have demonstrated that increased roughness on the aerofoil surface adversely affects efficiency \cite{Suder94}. Additionally, surface roughness has been found to influence flow separation topology, leading to increased hub end-wall losses due to three-dimensional suction side separation \cite{Gbadebo2004}. One popular method to model roughness analytically is to relate the centerline average roughness ($R_a$) to the equivalent sand grain roughness size ($k_s$). Various correlations have been proposed in the literature, with values ranging between $2 < k_s/R_a < 10$ \cite{Bons2010}. Specifying a constant roughness value has been shown to provide reasonable agreement with test data for manufactured blades, particularly in transonic compressors \cite{Kozulovic2022}. A parametric study to identify the most relevant correlation for gas turbine high Reynolds number applications \cite{Hummel2005}, resulted in the expression $k_s = 5.2 R_a$, which is utilised in this work.

\subsection{Machine Learning applications to turbomachinery}

In recent years, there has been a significant amount of research into the utilisation of machine learning techniques for turbomachinery applications. These include predicting the aerodynamic performance due to tip clearance variations \cite{Krishnababu2021} and design modifications \cite{Pongetti2021}, as well as aeromechanic predictions for both forced response \cite{Bruni2022} and flutter \cite{He2022}. As often the output variables are predicted directly from the input variables with a "black box" approach, these methodologies have varying degrees of accuracy and generalisability, depending on the complexity of the application. However, the ability to accurately predict variations in target variables for turbomachinery in a manner that is both reliable and generalisable, is contingent upon the accuracy of flow field predictions. It is only through precise flow field predictions that the overall performance figure of interest can be calculated through the application of relevant physical equations that govern the dynamics. The potential benefits of using machine learning approaches for predicting full flow fields, have been demonstrated in some simplified cases in the literature for 2D aerofoils on both cartesian \cite{Thuerey2020} \cite{chen2023} \cite{Ribeiro2020} and unstructured \cite{Kashefi2021} grids. CNNs have been demonstrated to predict the flow field downstream of a single fan row \cite{Li2023}, which was then used for noise predictions applications. A similar approach was used also by Rao et. al. \cite{Rao2023} for the semantic segmentation of an aero-engine intake using U-Net and U-Net++ architectures. Graph neural networks \cite{Harsch2021} \cite{Pfaff2021} have also been applied to 3D turbomachinery, with a single row compressor \cite{Perrone2022} and turbine \cite{Li2022}, which suggested potential challenges related to scalability and practicality for industrial applications. For instance, the computational cost associated with training, and the size of the required model, make the approach impractical for multi-stage compressors. These applications typically use computational meshes in the order of magnitude of 10 million nodes. Furthermore, storing the CFD data required for training and future simulations would require several petabytes, for a typical engine manufacturer, which is not practical. Purely data-driven approaches typical of supervised learning require large amount of data - so that the network can learn the solution manifold. A recent review \cite{thuerey2021pbdl} has highlighted the potential of other approaches such as Physics-Informed Neural Networks (PINNs), Reinforcement Learning (RL), and Generative Adversarial Networks (GANs) \cite{Chu2021} to mitigate these issues. However, none of those methods can currently provide accurate, generalisable and scalable solutions to complex 3D turbomachinery applications.

\section{Novel contribution: Physics-based dimensionality reduction} \label{sec:novelty}

To address these challenges, we introduce a pre-processing step to extract only the relevant engineering data from the 3D computational domain, such as: blade-to-blade planes, blade surfaces, axial cuts, loading distributions, radial distributions, stage-wise performance, and overall performance. In most cases, these provide sufficient information to the aerodynamicist for design and analyses activities, while close examination of the 3D flow field of a CFD solution is generally only necessary in non-standard cases. Once the relevant engineering data has been extracted from the computational domain, it can be interpolated onto a simplified grid with a resolution that is deemed acceptable for the application in question. The regression problem is therefore reformulated, from predicting the relevant variables on an unstructured computational grid in the order of magnitude of $\sim 10^7$ nodes, to a structured grid in the order of magnitude of $\sim 10^5$ nodes. Only selected fundamental variables are predicted, while the derived ones can be computed using the appropriate physical equations, significantly reducing the computational cost without sacrificing accuracy, as only the necessary data required to calculate the target variables is retained. Targeting the flow field predictions and using the relevant physical equations to calculate the corresponding overall performance, renders the methodology generalisable, while filtering only relevant parts of the CFD solution makes the methodology scalable to complex industrial applications. This physics-based dimensionality reduction, unlocks the potential to use supervised learning approaches for turbomachinery applications, as it re-formulates the regression problem from an un-structured to a structured one, as well as reducing the number of degrees of freedom. The authors first introduced this approach in a previous work for a single stage compressor application \cite{Bruni2023}, which is now extended to a multi-stage compressor application, with a computational domain one order of magnitude larger. The framework developed in the previous paper is demonstrated to be translated from an academic test case, to a complex real-world application with significant engineering relevance.  While the focus of the current work is on multi-stage axial compressors, the framework is directly applicable to any other turbomachinery application, such as axial turbines.

\section{Methodology}

A modern multi-stage industrial axial compressor consisting of 10 stages, as shown in \autoref{fig:domain_paper}, is considered in this work. Each stage is comprised of a row of rotor blades and a row of stator blades with a given blade count. The effect of tip clearance and surface roughness variations on each blade row are the focus of this study. Future work will include other manufacturing and build variations such as geometry variations, as well as different operating conditions. The extension of the model to a wider range of input features does not require in principle any alteration to the architecture and the framework, which has been developed to scale to an arbitrary number of variables.

\subsection{Data Generation}

\begin{figure*}[!htbp]
   \centering
      \includegraphics[width=\textwidth]{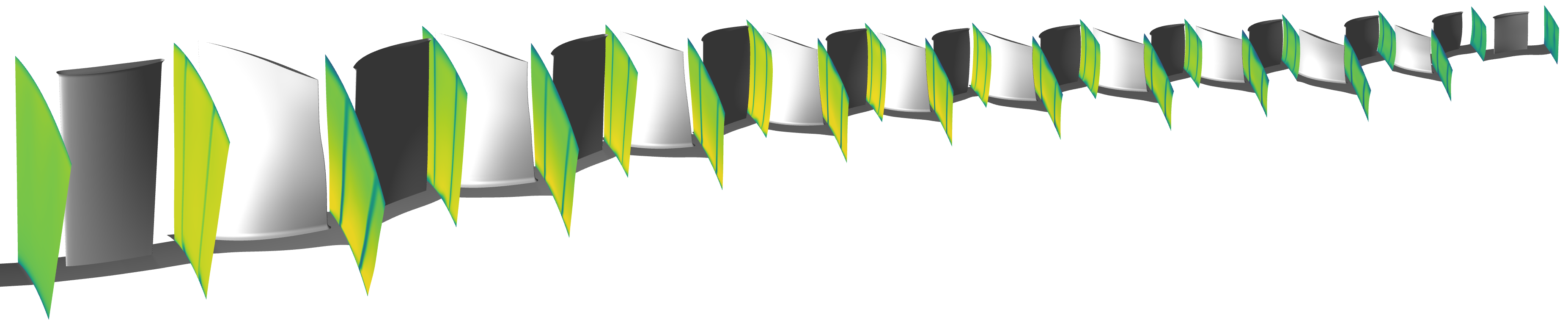}
      \caption{Overview of the CFD domain and axial velocity contours at the mixing-plane locations}
      \label{fig:domain_paper}
\end{figure*}

The ground-truth data used for training are the CFD results for the configuration of interest. This computational model consisted of 10 stages, with mixing plane interfaces. To reduce the computational cost of the CFD analysis, rotational periodicity is assumed, and only a single passage needs to be considered instead of the full annulus, as standard process for steady state CFD analysis. An overview of the computational domain is shown in \autoref{fig:domain_paper}, showing both the geometry of the airfoils as well as the axial velocity contours at the mixing plane locations. The computational mesh was generated using Numeca AUTOGRID5, and the CFD solver used was Trace, with SST $K-\omega$ turbulence model. More details on the computational setup used have been provided in previous publications \cite{Bruni2022}. The development of build-specific CFD models for compressor aerodynamic predictions and its validation against engine test data was also presented in previous work by the authors, discussing the relative effects of various manufacturing and build variations \cite{Bruni2024CFD}. The quantification of the accuracy of the CFD simulations in comparison to test data is outside of the scope of the current work. The aim of the proposed architecture is to provide predictions of the flow field and overall performance of the compressor with accuracy similar to that of the CFD model, but without the associated computational cost.

\begin{figure}[!htbp]
   \begin{subfigure}{0.49\textwidth}
   \centering
      \includegraphics[width=\textwidth]{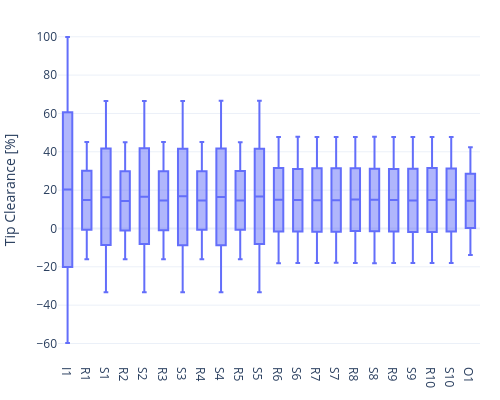}
   \end{subfigure}
   \begin{subfigure}{0.49\textwidth}
     \centering
        \includegraphics[width=\textwidth]{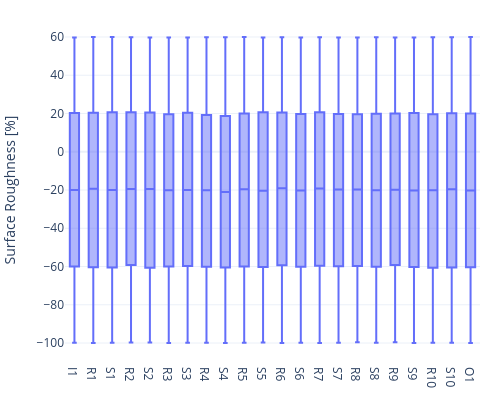}
     \end{subfigure}
     \caption{Box plots of the input variables (tip clearance and surface roughness) at each blade row}
     \label{fig:boxplot}
\end{figure}

The tip clearance and surface roughness were varied for each row (I1=Inlet Guide Vane, R1=Rotor-1, S1=Stator-1, etc.) within out-of-tolerance conditions up to 50\% larger and tighter than the drawing specifications, as shown in \autoref{fig:boxplot}. This range is significantly larger than the tolerance typically specified for gas turbine applications, and was selected to demonstrate the robustness of the methodology to out-of-tolerance cases. The input variable space was sampled using latin-hypercubic sampling.  The dataset generated for this study comprises 400 CFD solutions, each of which was executed using automated and parallelized processes, taking 90 minutes on 72 CPUs for each case. Even if the computational cost required to generate the dataset for supervised training is potentially high, it should be noted that most of the data is already available from internal work of the industrial partner. It is envisioned that the majority of the data used for training of this framework will be based on available data, with additional computations carried out only when required, to extend the dataset for wider generalisability. The use of a supervised learning approach is appropriate for industrial applications, as typically gas turbine manufacturers have access to a significant historical database of CFD calculations from previous analytical and design activities. In case of new applications for which historical data is not available, it's advisable to favour physics-informed models, as the computational cost associated with generating the data from scratch would render supervised models impractical.

\subsection{Data pre-processing}

As discussed in \autoref{sec:novelty}, the CFD results are pre-processed to reduce the dimensionality of the problem, without any loss of information. First, only six variables are extracted, namely, Total Pressure ($P_t$), Total Temperature ($T_t$), Axial Velocity ($V_x$), Tangential Velocity ($V_t$), Radial Velocity ($V_r$), and Density ($\rho$). The flow condition at each mesh node is fully described by the six variables and can be used to calculate all other relevant flow variables in the CFD solution. Second, only the 24 axial locations shown in \autoref{fig:domain_paper} are considered. Third, as the methodology aims to be generalisable, the data is interpolated for a given passage at a fixed circumferential location, irrespective of the periodic boundary definitions, number of points and discretization used in the mesh. As the location of the periodic boundaries and distribution of the nodes can vary depending on the meshing approach or geometry, this approach is not limited to a given mesh or geometry, as typical in most surrogate models developed for optimisations.  While the CFD mesh for each simulation comprises over 15 million nodes, the new pre-processed domain contains less than 100 thousand nodes overall, yet it includes all the necessary information for the desired engineering assessment. The new 3D flow field now consists now of only 24 axial 2D planes with a $64\times64$ grid. The data is stored as a tensor with a channel-first convention and a shape of $(6\times24\times64\times64)$, where the dimensions represent the number of variables, number of axial locations, radial nodes in the mesh, and tangential nodes in the mesh, respectively. Generally, 2D contours are only used for detailed analyses, whereas mass-flow averaging \cite{CumpstyAveraging} in the circumferential direction is more commonly used to obtain radial profiles of the variables of interest for most engineering applications. These radial profiles of each variable are then mass-flow averaged radially to obtain 1D averages, which are used to calculate stage-wise performance of the variables of interest, such as pressure-ratio ($PR$), and polytropic efficiency ($\eta_p$). In addition to the aerodynamic performance of the stage, it is possible to calculate the overall performance of the compressor. Predicting the stage-wise performance in addition to the overall performance is critical for implementation of the model in the gas turbine build process, as it provides actionable data on which blade rows should be targeted by corrective measures.

\subsection{The C(NN)FD framework}

The tip clearance and surface roughness values, as well as the relevant geometry design parameters of a specific build are fed as input to \textit{C(NN)FD}, which predicts the 3D flowfield, consisting of the 2D contours for all variables at the axial locations of interest. The outputs are mass-flow averaged to obtain the relevant radial profiles and 1D averages, which are then used to calculate stage-wise performance first, and then the overall performance using relevant thermodynamic equations. As the impact of tip clearance and surface roughness variations depends on the aerodynamic loading distribution of a given aerofoil, the methodology should be generalisable to different designs. Therefore, \textit{C(NN)FD} requires the geometries associated with each clearance value as input. This is achieved by considering a series of design parameters to describe the geometry of each blade, such as stagger angle, camber angle, maximum thickness, etc. Since we consider only a single compressor design, the geometry parameters are fixed. In future work we will investigate the effect of geometrical variations and different designs. 

\subsubsection{Feature Engineering}

The input of the network consists of two separate tensors. The first input is a tensor of size $(2\times22)$, where the dimensions represent the input variables for each airfoil, respectively tip clearance and surface roughness, and the number of blade rows in the compressor. The second input is a tensor of size $(5\times22\times8)$ with the blade geometry design parameters. The dimensions represent the number of design variables, number of rows and number of radial sections. The compressor consists of 22 blade-rows, and the aim of the network is to predict the flow field downstream of each of these blade-rows, at the corresponding 22 inter-row locations. However, for compressor aerodynamic analysis, often two additional locations are considered to calculate the overall performance: "Compressor Inlet" and "Compressor Outlet", respectively upstream of the first row and downstream of the last row. This brings the number of axial locations for the predictions to 24. Therefore, the input tensor is manipulated by duplicating the input data for the first row at the beginning of the tensor, and the input data for the last row at the end of the tensor, bringing the first dimension of the tensor to 24. The input geometry tensor includes 8 radial sections at which the design parameters are specified, but this number can vary depending on the aerodynamic design philosophy for the compressor design of interest. Therefore, the radial sections are interpolated to computational grid, bringing input geometry shape to $(5\times24\times64)$. The geometry definition is then copied in the tangential direction to match the computational grid, defining a tensor of shape $(5\times24\times64\times64)$. Likewise, the input tensor containing the tip clearance and surface roughness variables of shape $(2\times24)$ is copied in the radial and tangential direction, to give a tensor of shape  $(2\times24\times64\times64)$. This allows to map the 2D and 3D input tensors to the 4D output tensor required for the flow field predictions. The two tensors are then added, resulting in a $(7\times24\times64\times24)$ tensor, using a channel first convention. The number of channels is now representative of the input variables, including: tip clearance, surface roughness, inlet metal angle, outlet metal angle, maximum thickness, chord and pitch. The number of input variables could increase if considering more manufacturing and build variations. Therefore, a convolution is adopted to change the number of channels from the number of input design variables, to the number of output flow variables, which in this instance is 6. This results in a tensor of shape $(6\times24\times64\times64)$ matching the requirements for the output flow field predictions. If more variables were required for the flow field predictions, it would only be required to extract the relevant data from the CFD solution and to increase the number of channels accordingly. One advantage of this architecture compared to others in literature, is that all the 6 variables of interest are predicted by a single model, instead of having to re-train the model separately for each variable of interest.

\subsubsection{Neural Network architecture} \label{sec:network}

\begin{figure}[!htbp]
    \centering
       \includegraphics[width=\textwidth]{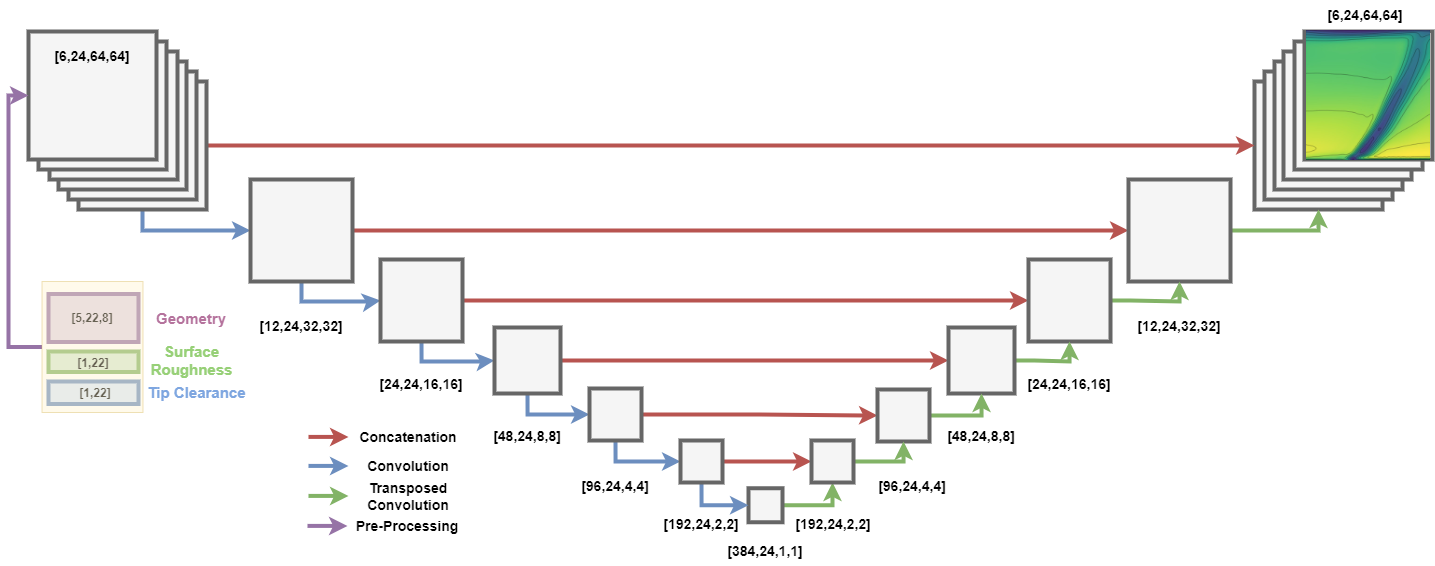}
       \caption{C(NN)FD architecture overview}
       \label{fig:architecture}
\end{figure}

 The architecture shown in \autoref{fig:architecture}, is a variant of the one originally proposed by the authors in previous work \cite{Bruni2023}. Following a pre-processing step for the input variables, the model is based on a 3D U-Net architecture with double residual convolutional blocks. Each convolutional block consists of a 3D convolution layer, followed by batch normalization and an \textit{ELU} activation function. A residual connection is implement from after the first convolution to before the second activation function. The use of residual connections improves convergence and makes this type of architecture scalable to larger models required for full compressor applications. The number of axial locations is fixed to 24, and each down-sampling convolutions and up-sampling transposed convolutions uses a stride of $(1,2,2)$, resulting in a bottleneck section with tensors of shape $(384\times24\times1\times1)$ after six layers. Both up and down samplings are followed by a double convolutional block at each layer. The output of the network is a tensor of size $(6\times24\times64\times64)$, which encompasses the entire flow field. The encoding section followed by the decoding section, along with the skip connections, enables the network to predict both low-level and high-level features in the flow field. 

\subsubsection{Training}

\begin{figure}[!htbp]
   \centering
      \includegraphics[width=0.8\textwidth]{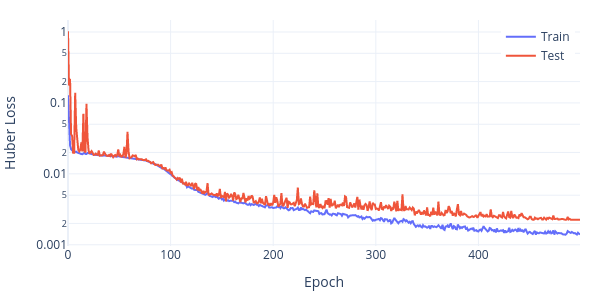}
      \caption{Training history: Training loss and Test loss}
      \label{fig:loss}
\end{figure}

The network is trained using the \textit{AdamW} optimizer, with \textit{Huber} loss function. The dataset was divided into three sets: 70\% training data, 20\% validation data, and 10\% hold-out data. The validation set was utilized to evaluate model performance and perform hyper-parameter tuning, while the hold-out set was reserved only for a final assessment. This ensures that the results are generalisable to unseen cases, and will lead to comparable performance also when deployed in a production environment. The dataset is split in a stratified manner, dividing the data in 10 bins using a quantile-based discretization function on the target overall efficiency. This ensures that each dataset has a comparable distribution. It should be noted that the dataset already contains out-of-tolerance cases and that no further out-of-distribution data outside of those limits is expected for the engineering application of interest. For further robustness, the training is performed using 5 different values of the random seed used for the dataset split, batch selection and initialization. The results are then presented providing the mean and the standard deviation of the losses. For instance, the final model showed a training loss of $(1.29 \pm 0.16) \cdot 10^{-3}$ and a test loss of $(1.96 \pm 0.40) \cdot 10^{-3}$ (Huber Loss $\pm$ standard deviation). An example of the training curve for one seed, starting from a random initialization in shown in \autoref{fig:loss}. The training runs generally converged within 500 epochs, with a batch size of 20 and an initial learning rate of 0.01. A learning rate scheduler was implemented, which halved the learning rate with a patience of 20 epochs. To avoid overfitting, early stopping with a patience of 50 epochs was implemented. The training time for a new model from scratch is around 1 hour on a single Tesla T4 GPU. This is reduced to less than 20 minutes when re-training the network when data becomes available, thus making the architecture scalable for industrial applications. The execution time for inference is less than 1 second, making the predictions of the deployed model effectively real-time for the application of interest.

\subsubsection{Model Comparison}
The novelty of the C(NN)FD framework has been discussed in Section \ref{sec:novelty}, highlighting how the appropriate pre-processing of the CFD solution and the usage of the relevant physical equations allow for the usage of convolutional neural networks for flow field predictions of complex 3D turbomachinery flows. Regardless, the complexity of the regression problem will increase with the number of input features considered, due for example to: number of stages in the compressor and associated geometry parametrization, manufacturing variations considered (tip clearance, surface roughness, geometry variations or more) and operating conditions. Therefore, the complexity and dimensionality of the regression problem distinguishes this work from previous literature \cite{Bruni2023}. For comparison, three different machine learning models are considered. The first one, \textit{Double Convolution} consists only two convolution layers, each of which with 3D convolution, batch normalization, and ELU activation function. This is considered just as a reference for what the simplest architecture that would be appropriate for the problem of interest. The second model, \textit{U-Net} is a 3D implementation of the standard U-Net, comparable to the ones used in previous literature for 2D airfoil flow predictions \cite{Thuerey2020}, and re-implemented for a 3D formulation.  The \textit{C(NN)FD} is the architecture presented in this paper, and discussed in more details in Section \ref{sec:network}. The main difference between the last two is that \textit{C(NN)FD} includes residual connections within the double convolution blocks. The use of residual connections \cite{ResUnet} was found to significantly improve the training for deeper networks, as it facilitates the propagation of the information across the network. Moreover, standard 3D \textit{U-Nets} use a constant stride of $(2,2,2)$ across the three dimensions when downsampling and upsampling, which due to the flow-field target dimension of $(6\times24\times64\times64)$ limits the width of the U-Net to three layers. In \textit{C(NN)FD} the number of axial locations is instead fixed to 24 by using a stride of $(1,2,2)$, which allows for a wider U-Net with a total of six layers. In addition to the more complex architecture, fixing the first dimension to $24$ as the number of axial locations, is physically more meaningful. The network can learn the complex relationships between the different inter-row locations, as the flow field at a given location is defined by the upstream and downstream conditions, as well as the input features representative of that blade-row (i.e. geometry parameters, tip clearance and surface roughness). The test loss of \textit{Double Convolution} is $10.3x$ higher than \textit{C(NN)FD}, while for the standard \textit{U-Net} is $3.28x$ higher. The accuracy achieved by these two architectures would not be sufficient to be used for the engineering application of interest. The improved accuracy of \textit{C(NN)FD} is due to the residual connections, more physically meaningful convolutions, and higher number of trainable parameters for \textit{C(NN)FD}. However, more complex models such as U-Net++ \cite{Zhou2018} did not lead to a significant improvement in the accuracy of the predictions, and were not considered worth of the increase computational cost. Regardless, they will be considered in future work for more complex datasets, such as the ones including also geometrical variations or different operating conditions.

\begin{table}[!htbp]
   \centering    
   \caption{Model comparison and benchmarking for different CNN architectures}
   \begin{tabular}{c|cccc}
    Model              & Trainable Parameters &  Wall Time [h] & Train Loss & Test Loss    \\ \hline
    \textit{Double Convolution} & $5.22\cdot 10^{3}$   & $0.8$          & $17.39\cdot 10^{-3}$    & $17.85\cdot 10^{-3}$    \\ 
    \textit{U-Net}        & $3.19\cdot 10^{5}$   & $2.3$          & $5.40\cdot 10^{-3}$     & $5.67\cdot 10^{-3}$     \\ 
    \textit{C(NN)FD}            & $1.28\cdot 10^{7}$   & $3.6$          & $1.26\cdot 10^{-3}$     & $1.73\cdot 10^{-3}$     \\ 
   \end{tabular}
\label{table:OverallPerformance_cmp}
\end{table}

\section{Results}

In this section, the predicted flow field generated by \textit{C(NN)FD} is compared to the CFD ground truth for the worst case scenario, where the largest discrepancy between the two was found in the hold-out set. Rather than selecting a cherry-picked case, as often the case in literature, comparing the results for the worst performer allows to identify the shortcoming of the model with regards to their physical interpretation and understand which flow features are more challenging to model using supervised learning approaches. 

\subsection{Flow field predictions}
\begin{figure}[!htbp]
   \begin{subfigure}{0.63\textwidth}
   \centering
      \includegraphics[width=\textwidth]{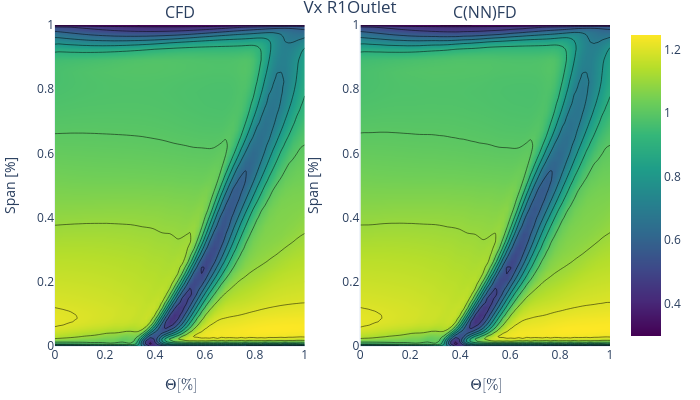}
   \end{subfigure}
   \begin{subfigure}{0.37\textwidth}
     \centering
     \includegraphics[width=\textwidth]{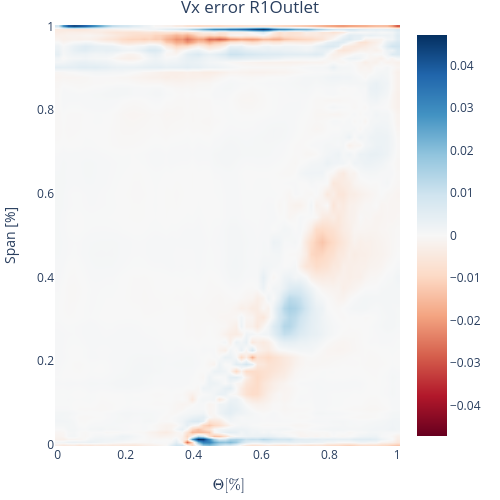}
     \end{subfigure}
     \caption{$V_x$ contour comparison between CFD and C(NN)FD predictions: Rotor 1 outlet}
     \label{fig:Worst_rotor_1}
\end{figure}

\begin{figure}[!htbp]
   \begin{subfigure}{0.63\textwidth}
   \centering
      \includegraphics[width=\textwidth]{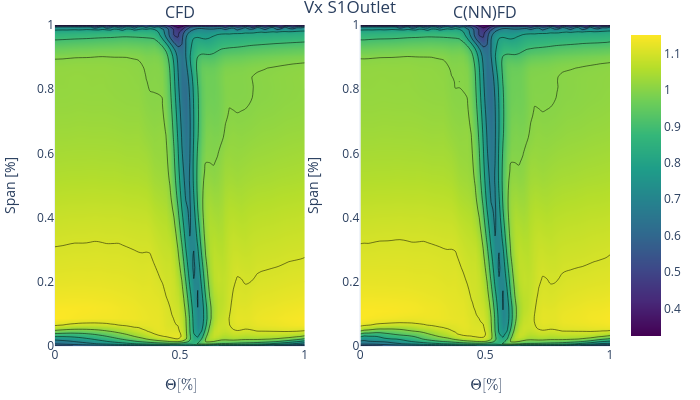}
   \end{subfigure}
   \begin{subfigure}{0.37\textwidth}
     \centering
     \includegraphics[width=\textwidth]{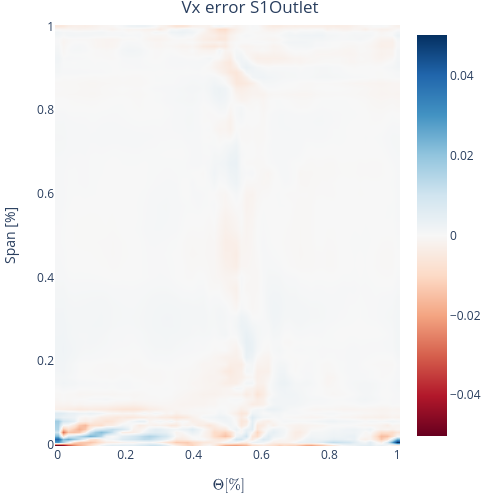}
     \end{subfigure}
     \caption{$V_x$ contour comparison between CFD and C(NN)FD predictions: Stator 1 outlet}
     \label{fig:Worst_stator_1}
\end{figure}

\begin{figure}[!htbp]
   \begin{subfigure}{0.63\textwidth}
   \centering
      \includegraphics[width=\textwidth]{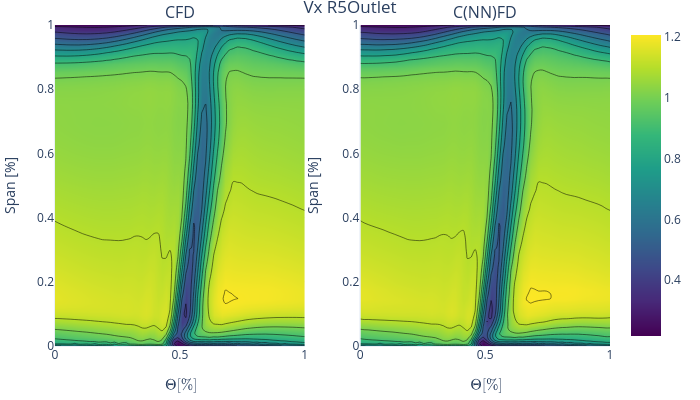}
   \end{subfigure}
   \begin{subfigure}{0.37\textwidth}
     \centering
     \includegraphics[width=\textwidth]{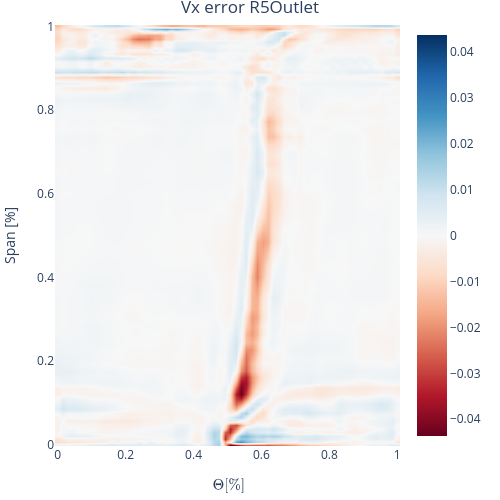}
     \end{subfigure}
     \caption{$V_x$ contour comparison between CFD and C(NN)FD predictions: Rotor 5 outlet}
     \label{fig:Worst_rotor_5}
\end{figure}

\begin{figure}[!htbp]
   \begin{subfigure}{0.63\textwidth}
   \centering
      \includegraphics[width=\textwidth]{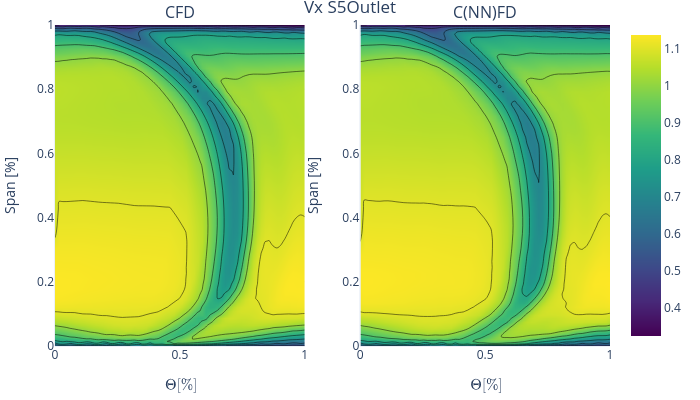}
   \end{subfigure}
   \begin{subfigure}{0.37\textwidth}
     \centering
     \includegraphics[width=\textwidth]{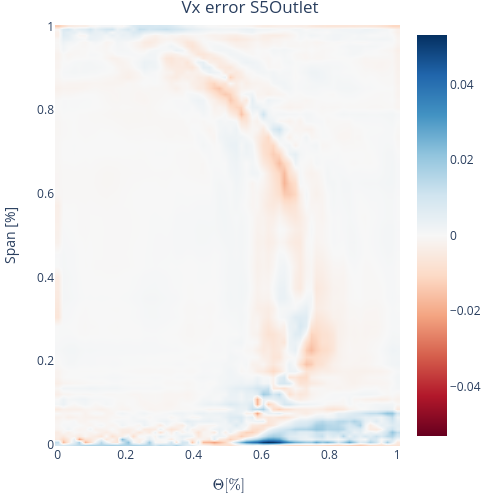}
     \end{subfigure}
     \caption{$V_x$ contour comparison between CFD and C(NN)FD predictions: Stator 5 outlet}
     \label{fig:Worst_stator_5}
\end{figure}

\begin{figure}[!htbp]
   \begin{subfigure}{0.63\textwidth}
   \centering
      \includegraphics[width=\textwidth]{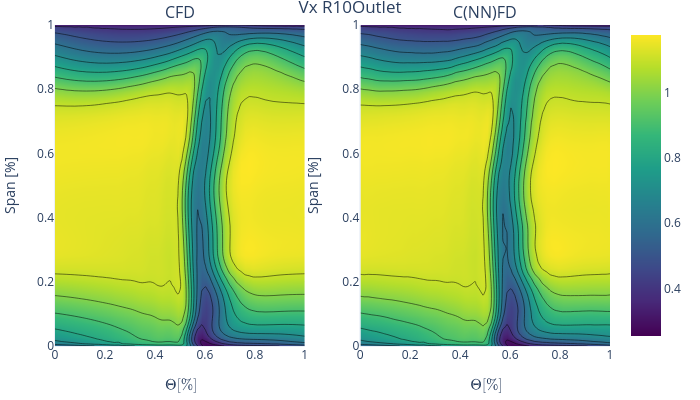}
   \end{subfigure}
   \begin{subfigure}{0.37\textwidth}
     \centering
     \includegraphics[width=\textwidth]{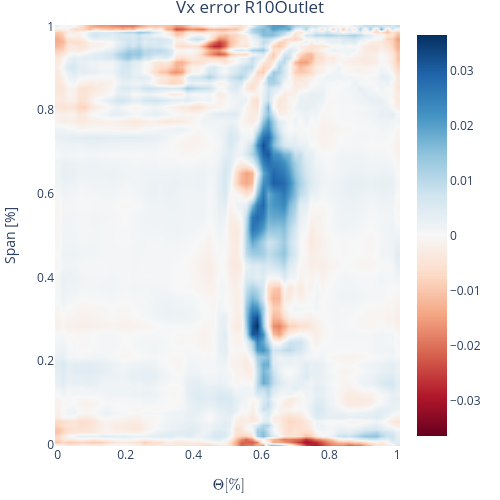}
     \end{subfigure}
     \caption{$V_x$ contour comparison between CFD and C(NN)FD predictions: Rotor 10 outlet}
     \label{fig:Worst_rotor_10}
\end{figure}

\begin{figure}[!htbp]
   \begin{subfigure}{0.63\textwidth}
   \centering
      \includegraphics[width=\textwidth]{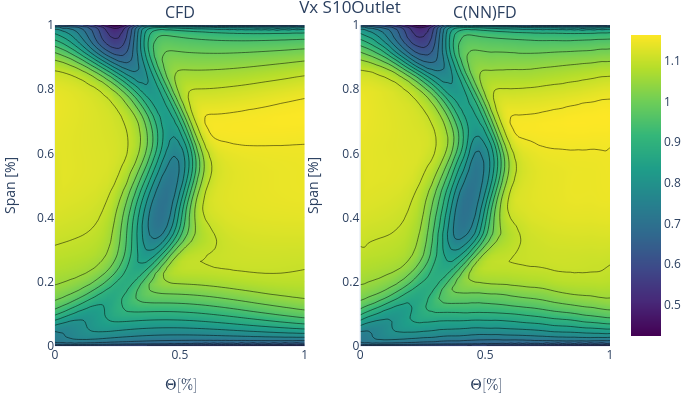}
   \end{subfigure}
   \begin{subfigure}{0.37\textwidth}
     \centering
     \includegraphics[width=\textwidth]{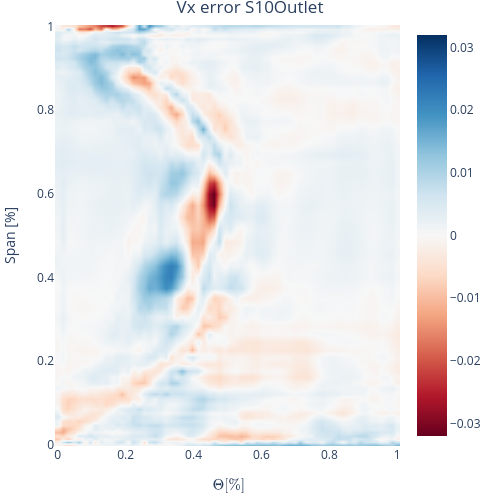}
     \end{subfigure}
     \caption{$V_x$ contour comparison between CFD and C(NN)FD predictions: Stator 10 outlet}
     \label{fig:Worst_stator_10}
\end{figure}

The results are presented for Axial Velocity $V_x$ in \autoref{fig:Worst_rotor_1}, \autoref{fig:Worst_stator_1}, \autoref{fig:Worst_rotor_5}, \autoref{fig:Worst_stator_5} and \autoref{fig:Worst_rotor_10}, \autoref{fig:Worst_stator_10} at Stage-1, Stage-5 and Stage-10, with CFD predictions in the left, C(NN)FD predictions in the middle, and relative error between the ground truth and the predictions at each mesh node on the right. All the other stages and variables exhibit similar level of agreement and are not presented for conciseness. The effect of errors in the other variables (Total Pressure ($P_t$), Total Temperature ($T_t$), Tangential Velocity ($V_t$), Radial Velocity ($V_r$), and Density ($\rho$)) are embedded in the assesment of the stage-wise distributions and overall performance, as they are used to calculate them directly through mass-flow averaging and the relevant physical equations. The $V_x$ comparison presented is used to judge the quality of the machine learning model in predicting the relavant aerodynamic flow features. This includes for example the regions of low axial momentum associated with the airfoil wakes, the tip clearance flows and suction side separations. All values presented are non-dimensionalized with respect to the results obtained from the baseline. The \textit{Span [\%]} y-axis represents the radial direction in the gas path, ranging from the hub surface at the bottom to the casing surface at the top. The $\theta [\%]$ x-axis represents the circumferential direction in the gas path, with a range between 0 and 1 representing one passage. Each row corresponds to a specific number of blades, and thus the gas path can be divided into a number of passages equal to the blade count. The predictions generated by \textit{C(NN)FD} exhibit excellent agreement with the ground truth, with the primary flow features being accurately reproduced. Only minor differences are observed in regions with high gradients, but well within the numerical uncertainty of the CFD solver. Due to operational requirements, the aerodynamic design of axial compressor leads to greatly different flow features across the different stages. For instance, front stages tend to operate at transonic conditions, with strong shock-waves interacting with the tip leakage flows. Conversely, the rear stages are subsonic, with the flow field dominated by end-wall flow features. The proposed architecture is demonstrated to be able to predict the flow field within the compressor for both transonic and subsonic stages, even with significantly different flow features as seen for instance by comparing \autoref{fig:Worst_rotor_1} and \autoref{fig:Worst_rotor_10}. The highest local error is less than $0.05\%$, which is well within the numerical error of the CFD solver. The errors are generally localised to regions with steep gradients and to only few nodes in the mesh, being therefore effectively negligible from an engineering perspective.


\newpage
\subsection{Derived variables predictions}

\begin{figure}[!htbp]
    \begin{subfigure}{0.49\textwidth}
    \centering
       \includegraphics[width=\textwidth]{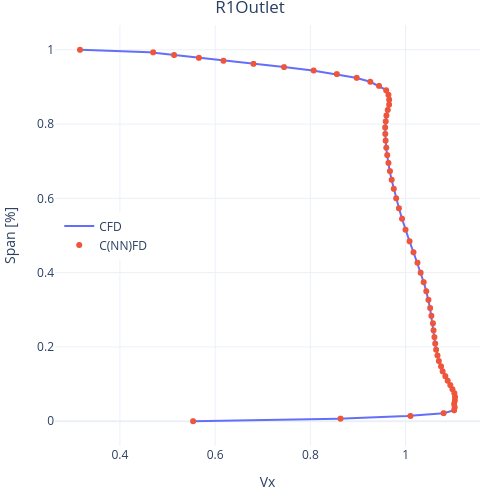}
    \end{subfigure}
    \begin{subfigure}{0.49\textwidth}
      \centering
         \includegraphics[width=\textwidth]{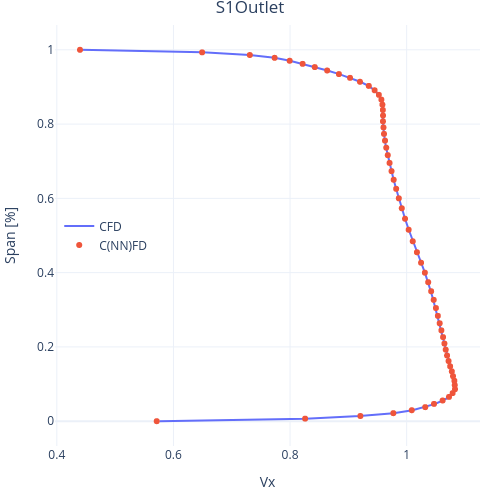}
      \end{subfigure}
      \caption{Radial profiles comparison between CFD and C(NN)FD: Stage 1 - Rotor and Stator outlet}
      \label{fig:Worst_Radial_1}
\end{figure}

\begin{figure}[!htbp]
   \begin{subfigure}{0.49\textwidth}
    \centering
       \includegraphics[width=\textwidth]{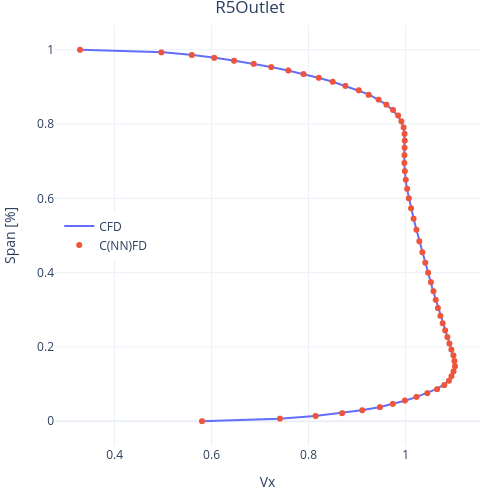}
    \end{subfigure}
    \begin{subfigure}{0.49\textwidth}
      \centering
         \includegraphics[width=\textwidth]{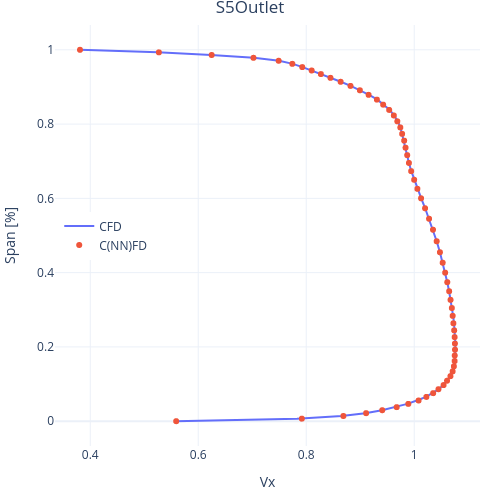}
      \end{subfigure}
      \caption{Radial profiles comparison between CFD and C(NN)FD: Stage 5 - Rotor and Stator outlet}
      \label{fig:Worst_Radial_5}
\end{figure}

\begin{figure}[!htbp]
    \begin{subfigure}{0.49\textwidth}
    \centering
       \includegraphics[width=\textwidth]{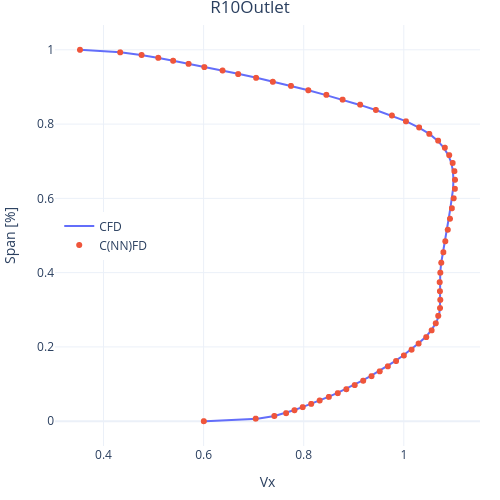}
    \end{subfigure}
    \begin{subfigure}{0.49\textwidth}
      \centering
         \includegraphics[width=\textwidth]{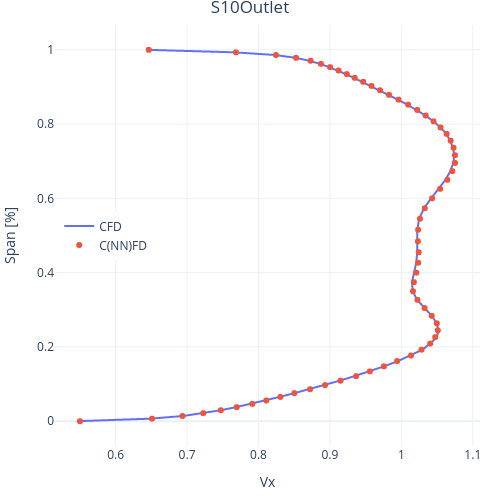}
      \end{subfigure}
      \caption{Radial profiles comparison between CFD and C(NN)FD: Stage 10 - Rotor and Stator outlet}
      \label{fig:Worst_Radial_10}
\end{figure}

While C(NN)FD is trained exclusively on the flow field results from CFD, the 2D contours predictions are also used as a stepping stone to calculate other variables that are more frequently used for the engineering assessments. Mass-flow averaging in the circumferential direction is used to obtain radial profiles of the variables, which are then mass-flow averaged to obtain a 1D average, which is used to calculate stage-wise performance and overall performance of the variables of interest, such as pressure-ratio ($PR$), and polytropic efficiency ($\eta_p$). When performing mass-flow averaging to obtain the radial profiles, the discrepancies discussed for the 2D contours between the ground truth and predictions are even less significant. \autoref{fig:Worst_Radial_1}, \autoref{fig:Worst_Radial_5} and \autoref{fig:Worst_Radial_10} indicate that the radial profiles for \textit{C(NN)FD} and CFD are effectively overlapping. The results are presented for Stage-1,  Stage-5 and Stage-10. All the other stages and variables exhibit similar level of agreement and are not presented for conciseness. It is notable how even the end-wall regions, characterized by steep gradients, are well-resolved by \textit{C(NN)FD}. This confirms that the localized errors found in the 2D contours prediction do not significantly affect the results for the radial profiles. The rationale for including the radial profiles in the framework, is to allow the aerodynamic engineer to identify the flow features driving a change in performance, in a more actionable form than 2D contours. For instance, larger tip clearance could lead to more blockage due to tip flows in the casing region, resulting in lower axial momentum and increased mixing losses. This would then be apparent in the radial profiles, that could allow to explain the corresponding change in stage matching and overall performance. 


\begin{figure}[!htbp]
   \centering
   \begin{subfigure}{0.49\textwidth}
      \includegraphics[width=\textwidth]{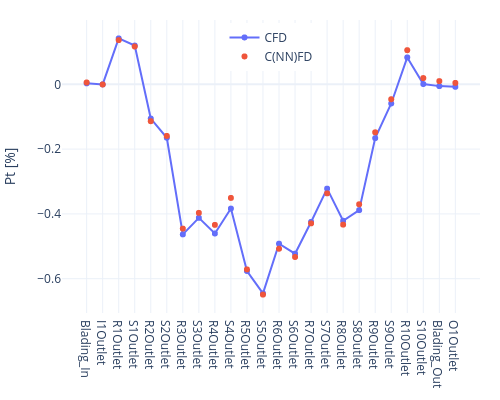}
      \caption{$P_t$ row-wise distribution}
      \label{fig:row_wise_Pt}
   \end{subfigure}
   \begin{subfigure}{0.49\textwidth}
   \centering
      \includegraphics[width=\textwidth]{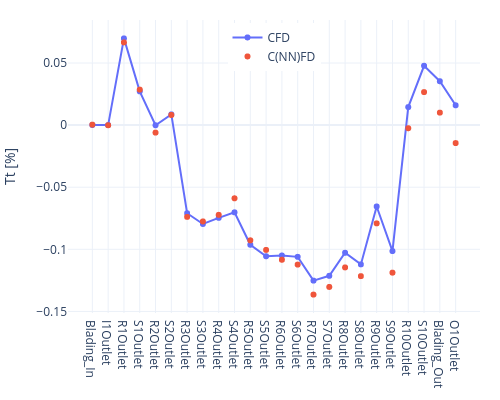}
      \caption{$T_t$ row-wise distribution}
      \label{fig:row_wise_Tt}
   \end{subfigure}
   \centering
   \begin{subfigure}{0.49\textwidth}
      \includegraphics[width=\textwidth]{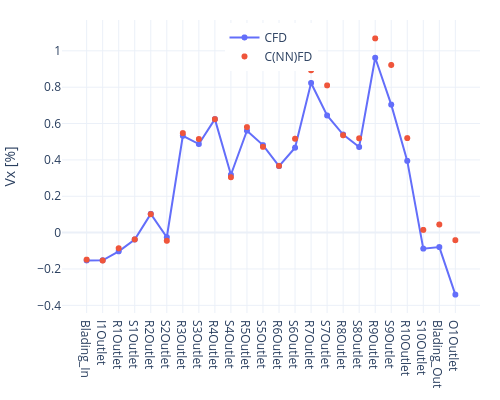}
      \caption{$V_x$ row-wise distribution}
      \label{fig:row_wise_Vx}
   \end{subfigure}
   \begin{subfigure}{0.49\textwidth}
   \centering
      \includegraphics[width=\textwidth]{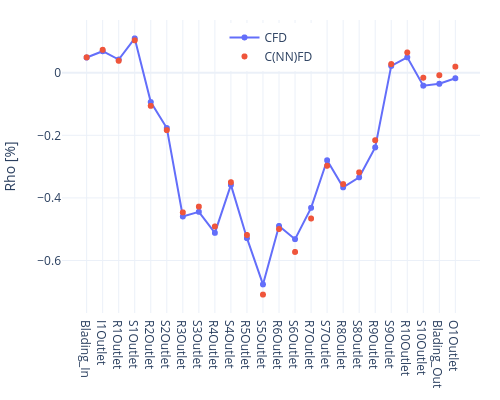}
      \caption{$\rho$ row-wise distribution}
      \label{fig:row_wise_rho}
   \end{subfigure}
   \caption{Row-wise distributions comparison between predicted (C(NN)FD) and ground truth (CFD) values for worst case in the hold-out set}
\end{figure}

\newpage
The radial profiles are then mass-flow averaged to obtain the relevant 1D averages. An overview of the 1D average at each inter-row location is shown for Total Pressure $P_t$ in \autoref{fig:row_wise_Pt}, Total Temperature $T_t$ in \autoref{fig:row_wise_Tt}, Axial Velocity $V_x$ in \autoref{fig:row_wise_Vx} and Density $\rho$ in \autoref{fig:row_wise_rho}. Only these variables are presented as required to calculate the stage-wise and overall performance. However, similar agreement was also found for the remaining variables. Generally the trends are well captured, with $P_t$ and $\rho$ showing only negligible differences compared to the CFD predictions. While errors in $T_t$ and $V_x$ are slightly more noticeable for the rear stages of the compressor, they are still within the numerical uncertainty of the CFD calculations. Interestingly, while small in absolute terms, the errors are found to progressively increase towards the rear stages of the compressor. This is physically justifiable, as the flow field at these locations will be heavily influenced by the upstream conditions. Therefore, negligible errors in the front stages will progressively propagate towards the rear stages, leading to less accurate predictions, due to the higher uncertainty associated with the upstream flow field conditions. Future work will focus on quantifying the uncertainty of the predictions at each inter-row location, establishing the relative source of uncertainty which could be due to either shortcomings of the modelling of the effect of the input features for a given blade row, or the effect of the upstream and downstream boundary conditions for a given axial plane. It should be noted that mass-flow conservation at each inter-row location is critical in ensuring that the predictions from the framework are physically accurate. While the error in $\rho$ and $V_x$ is small, as presented in \autoref{fig:row_wise_rho} and \autoref{fig:row_wise_Vx}, mass-flow conservation is not set as a constraint during the training process. Future work will explore including mass conservation in the loss function.

\begin{figure}[!htbp]
   \centering
   \begin{subfigure}{0.49\textwidth}
      \includegraphics[width=\textwidth]{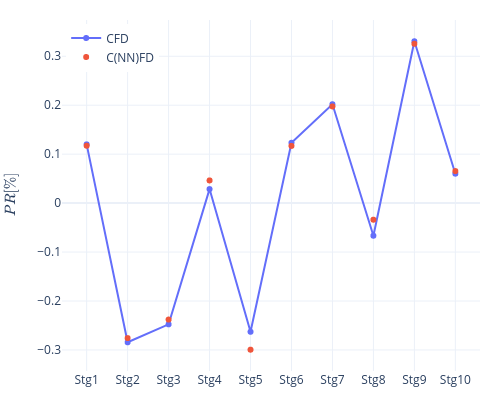}
      \caption{PR Stage-wise distribution}
      \label{fig:Stagewise_PR}
   \end{subfigure}
   \begin{subfigure}{0.49\textwidth}
   \centering
      \includegraphics[width=\textwidth]{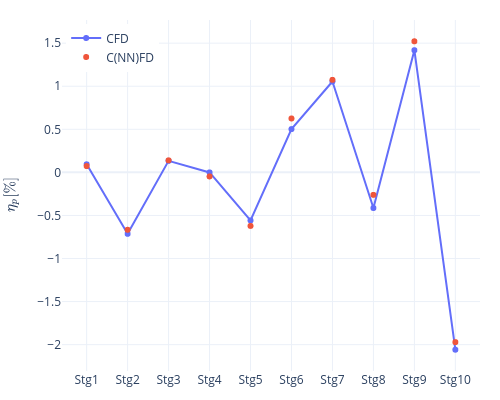}
      \caption{$\eta_p$ Stage-wise distribution}
      \label{fig:Stagewise_EtaP}
   \end{subfigure}
   \caption{Comparison between predicted (C(NN)FD) and ground truth (CFD) values for worst case in the hold-out set}
\end{figure}

\newpage
The 1D averages are then utilized to compute the stage-wise performance as shown in \autoref{fig:Stagewise_PR} for pressure ratio and \autoref{fig:Stagewise_EtaP} for polytropic efficiency. While the agreement is excellent, minor discrepancies between predictions and ground truth are noticeable. C(NN)FD predicts accurately the trend for each stage, relative to the baseline, with the variations being slightly over-predicted or under-predicted depending on the stage but being ultimately negligible.

\begin{figure}[!htbp]
    \begin{subfigure}{0.49\textwidth}
    \centering
       \includegraphics[width=\textwidth]{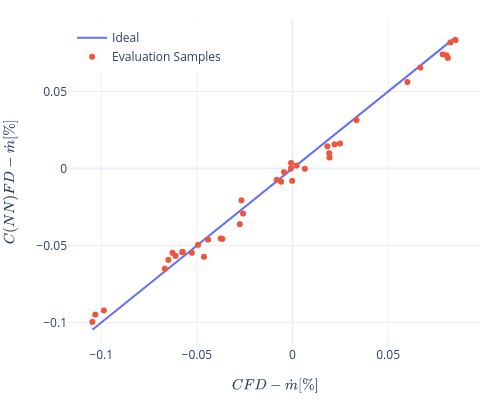}
       \caption{Mass-flow}
       \label{fig:Overall_Mass}
    \end{subfigure}
    \begin{subfigure}{0.49\textwidth}
    \centering
       \includegraphics[width=\textwidth]{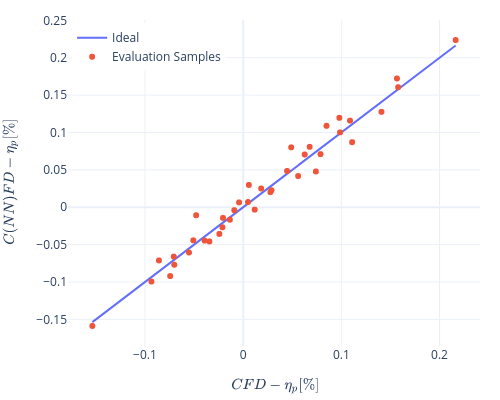}
       \caption{Polytropic Efficiency}
       \label{fig:Overall_EtaP}
    \end{subfigure}
    \caption{Comparison between predicted (C(NN)FD) and ground truth (CFD) values}
\end{figure}

Finally, the overall performance predictions are presented in \autoref{table:OverallPerformance}, \autoref{fig:Overall_Mass} and \autoref{fig:Overall_EtaP}. An excellent agreement is observed for all overall performance parameters, with a coefficient of determination $R^2$ close to 1 for all variables. This indicates that the proposed model can accurately describe most of the variance present in the dataset. Moreover, the Mean-Absolute-Error for each variable is smaller than the known uncertainties of the CFD ground truth results and significantly smaller than the dataset's range. If required, the predictions could potentially be further improved by providing training data including more extreme conditions, where it was deemed relevant from an industrial perspective.

\begin{table}[!htbp]
   \centering    
   \caption{Overall performance comparison}
   \begin{tabular}{c|ccc}
    
    Var       & $R^2$ & $MAE \%$   & $\Delta dataset \%$   \\ \hline
    $\dot{m}$ & 0.987 & 0.004          &  0.25              \\      $\eta_p$  & 0.968 & 0.011          &  0.47              \\ 
   \end{tabular}
\label{table:OverallPerformance}
\end{table}

\newpage
The \textit{C(NN)FD} predictions remain accurate even for the case with the lowest efficiency in the hold-out set, demonstrating the model's capability to handle challenging aerodynamic conditions typical of large clearances and surface roughness. The largest discrepancy shows a difference of less than $0.05$ percentage points in overall efficiency. This discrepancy is negligible for overall performance predictions, and demonstrates how the proposed architecture reliably predicts even the most challenging cases. This is a particularly challenging case aerodynamically, as high and low efficiency and pressure ratio stages are alternated, as shown in \autoref{fig:Stagewise_EtaP} . This would result for example in strong tip leakage flows leading to high losses and low efficiency in the blade rows with large clearances. Those would affect the downstream blade rows with tight clearance, whose potentially high efficiency will be affected by the unfavorable upstream conditions, leading to complex inter-row interactions and stage re-matching. This demonstrates how C(NN)FD can be used as a predictive tool to assess the performance of a given engine build, without the need for computationally expensive CFD analysis, and potentially avoiding re-testing for engines which would not meet the contractual performance requirements. Moreover, this tool can be used for a selective build process, where different parts available in stock can be selected based on the expected overall performance. The focus of this work was on compressor performance at design points conditions, while future work will extend the assessment to more off-design conditions and surge margin predictions. 

\subsection{Framework explainability}

Predicting the flow field at all the inter-row axial locations, as well as radial profiles, 1D averages and stage-wise performance in addition to the overall performance is critical for implementation of the model in the gas turbine build process. The key advantage is that the wealth of information available provides actionable data on which blade rows should be targeted by corrective measures. The proposed architecture has a significant competitive advantage compared to traditional surrogate models, which would directly predict the overall performance of the compressor, starting from a defined set of input variables. In addition to being less generalisable and requiring more data for training, traditional "black-box" models have the disadvantage of lacking explainability. Once a given overall performance variation is identified, it's possible to back trace the results to the stage-wise distributions, 1D averages, radial profiles and ultimately 2D contours. This would be of significant use for root cause analysis, in which a variation in performance due to a certain input features, can be addressed by identifying the aerodynamic drivers associated with it. Further applications would consist of surrogate models for design activities or optimisations, once the model is trained considering also geometrical variations. 

\section{Conclusion}

This paper demonstrates the development and application of a deep learning framework for predictions of the flow field and aerodynamic performance of multi-stage axial compressors. A physics-based dimensionality reduction unlocks the potential to use supervised learning approaches for flow-field predictions, as it re-formulates the regression problem from an un-structured to a structured one, as well as reducing the number of degrees of freedom. The proposed architecture also has a competitive advantage compared to traditional "black-box" surrogate models, as it provides explainability to the predictions of variations in overall performance by identifying the corresponding aerodynamic drivers. This is applied to model the effect of manufacturing and build variations, as the associated scatter in efficiency is known to have a significant impact on the overall performance and $CO_2$ emissions of the gas turbine, therefore posing a challenge of great industrial and environmental relevance. The proposed architecture is proven to achieve an accuracy comparable to that of the CFD benchmark, in real-time, for an industrially relevant application. The deployed model, is readily integrated within the manufacturing and build process of gas turbines, thus providing the opportunity to analytically assess the impact on performance with actionable and explainable data.

\begin{Backmatter}

\paragraph{Acknowledgments}
The authors would like to thank Siemens Energy Industrial Turbomachinery Ltd. for allowing the publication of this research, as well as Richard Bluck and Roger Wells for their support and comments. 

\paragraph{Funding Statement}
This work was supported and funded by Siemens Energy Industrial Turbomachinery Ltd.

\paragraph{Competing Interests}
Giuseppe Bruni and Senthil Krishnababu are currently employed by Siemens Energy Industrial Turbomachinery Ltd.

\paragraph{Data Availability Statement}
The data used is confidential and propriety of Siemens Energy Industrial Turbomachinery Ltd., and cannot be made available. 

\paragraph{Ethical Standards}
The research meets all ethical guidelines, including adherence to the legal requirements of the study country.

\paragraph{Author Contributions}
Conceptualization: G.B; S.M; S.K.  Methodology: G.B; S.M. Data curation: G.B. Data visualisation: G.B. Supervision: S.M. Funding acquisition: S.K. Writing - original draft: G.B. Writing - review $\&$ editing: G.B; S.M; S.K. All authors approved the final submitted draft.

\paragraph{Supplementary Material}
No supplementary material intended for publication has been provided with the submission.

\bibliographystyle{plainnat}
\bibliography{cnnfd_fc}

\end{Backmatter}

\end{document}